%% file: Schmerling.Pavone.RSS17.ARXIV.tex
\documentclass[conference]{IEEEtran}
\usepackage{times}

\IEEEoverridecommandlockouts                              % This command is only needed if 
\newif\ifarxiv

\newif\ifimages
\imagestrue

\arxivtrue

\usepackage{makeidx}         % allows index generation
\usepackage{graphicx}        % standard LaTeX graphics tool
\ifarxiv
\fi

\makeindex             % used for the subject index
\usepackage[numbers]{natbib}
\usepackage{multicol}
\usepackage[bookmarks=true]{hyperref}
\usepackage{mathtools}

\usepackage{enumerate}
\usepackage{epstopdf}
\usepackage{subfigure}
\usepackage{cleveref}

\usepackage{tablefootnote}

\usepackage{amsmath,amssymb}%,amsthm}
\usepackage{algorithm}
\usepackage{algorithmic}

\usepackage[usenames,dvipsnames,svgnames]{xcolor}

\graphicspath{{./fig/}}
\input{preamble}

\newcommand{\prob}[1]{\mathbb{P}\left( #1 \right) }

\renewcommand{\u}{\mathbf{u}}

\renewcommand{\v}{\mathbf{v}}

\newcommand{\mub}{\boldsymbol{\mu}}

\renewcommand{\baselinestretch}{.966}
\title{\LARGE \bf
Evaluating Trajectory Collision Probability through\\ Adaptive Importance Sampling for Safe Motion Planning
}

\author{\authorblockN{Edward Schmerling}
\authorblockA{Institute for Computational and Mathematical Engineering\\
Stanford University,
Stanford, CA 94305\\
Email: schmrlng@stanford.edu\vspace{-0.5cm}}
\and
\authorblockN{Marco Pavone}
\authorblockA{Department of Aeronautics and Astronautics\\
Stanford University,
Stanford, CA 94305\\
Email: pavone@stanford.edu\vspace{-0.5cm}}}

\begin{document}

\maketitle 

\begin{abstract}
This paper presents a tool for addressing a key component in many algorithms for planning robot trajectories under uncertainty:
evaluation of the safety of a robot whose actions are governed by a closed-loop feedback policy near a nominal planned trajectory.
We describe an adaptive importance sampling Monte Carlo framework that enables the evaluation of a given control policy for
satisfaction of a probabilistic collision avoidance constraint which also provides an associated certificate of accuracy (in the form of a confidence interval). In
particular this adaptive technique is well-suited to addressing the complexities of rigid-body collision checking applied to non-linear robot dynamics. As a Monte Carlo method
it is amenable to parallelization for computational tractability, and is generally applicable to a wide gamut of simulatable
systems, including alternative noise models. Numerical experiments demonstrating the effectiveness of the adaptive importance
sampling procedure are presented and discussed.
\end{abstract}

\IEEEpeerreviewmaketitle

\section{Introduction} \label{sec:intro}

The ability to anticipate the future and quantify its uncertainty is one of the most important features to enable safe and efficient decision making for autonomous robotic systems. While in deterministic problem settings a robot may simply base its decision making on unique projections of itself and the surrounding world, in uncertain settings a robot needs to either directly forward-simulate the many ways in which the future may evolve, or leverage analytically-derived heuristics/bounds that otherwise account for these many possibilities. In this paper we focus on the former approach, i.e., Monte Carlo (MC) simulation, which has been successfully applied to such complex tasks as visual feature selection for odometry \cite{CarloneKaraman2017}, online closed-loop belief state planning \cite{LuoBaiEtAl2016}, and planning safe kinodynamic trajectories subject to sensing and actuation uncertainty \cite{JansonSchmerlingEtAl2015b}. The strength of MC methods is that they need not make any over-simplifying or over-conservative assumptions in order to accommodate higher levels of detail (their accuracy is limited only by simulation fidelity); their main weakness is that detailed simulation of many futures may be computationally expensive. To mitigate this expense, MC methods may bias their simulations towards rare, but critical events, a technique applied in \cite{JansonSchmerlingEtAl2015b,LuoBaiEtAl2016}, in order to more efficiently estimate important performance characteristics in a process known as \emph{importance sampling} (IS). The objective of this paper is to devise an importance sampling MC framework to assess the safety of a given motion control policy, with an associated certificate of accuracy, in general, non-linear problem settings.

\begin{figure}[t]
\centering
    \includegraphics[width=0.45\textwidth]{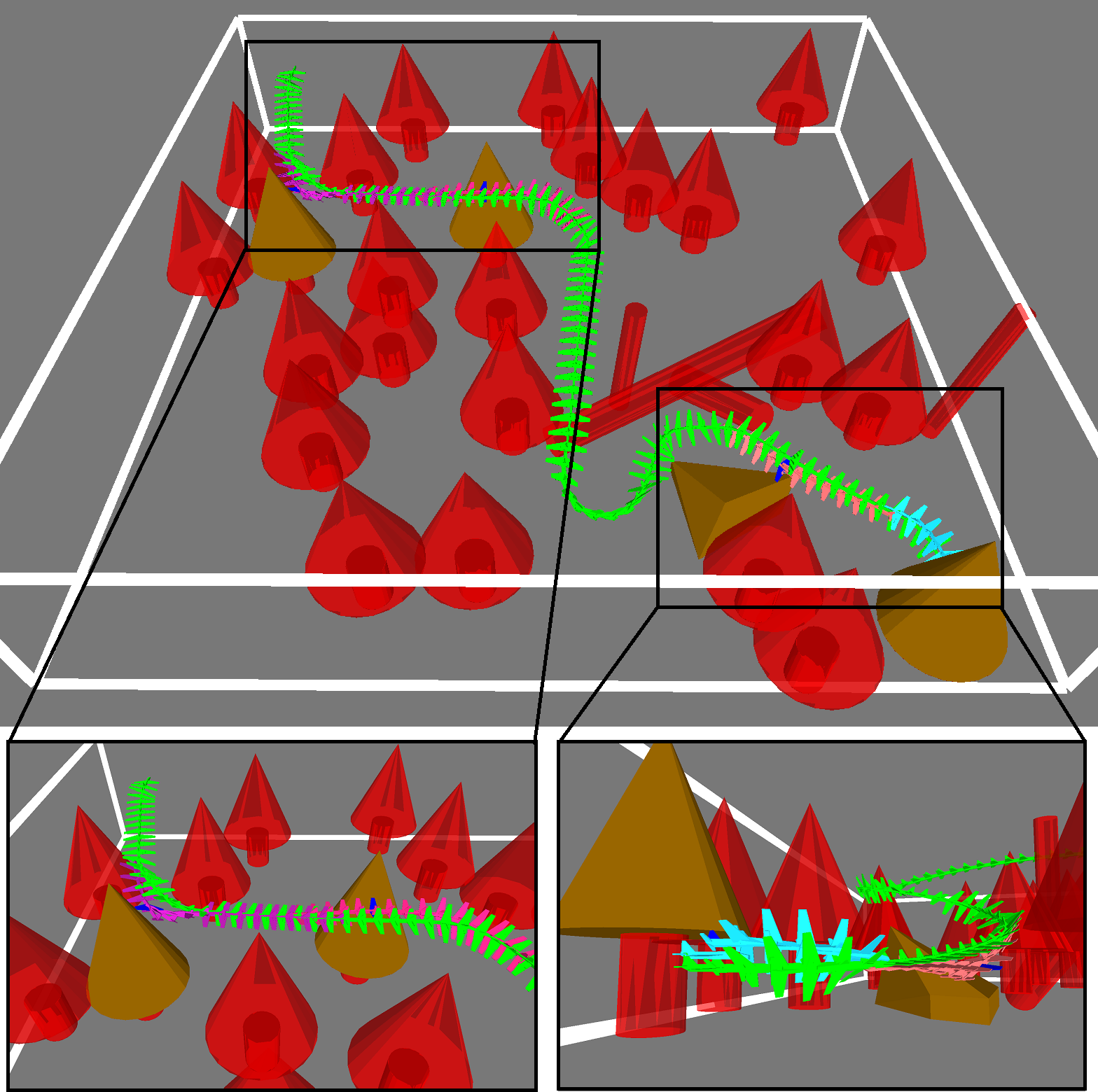}
    % \subfigure[]{\label{fig:full_crash} \includegraphics[width=0.44\textwidth]{plane_full.png}}
    % \subfigure[]{\label{fig:crash_zoom1} \includegraphics[width=0.22\textwidth]{plane_zoom1.png}}
    % \subfigure[]{\label{fig:crash_zoom2} \includegraphics[width=0.22\textwidth]{plane_zoom2.png}}
        \caption{Four likely collision modes for an airplane tracking a nominal trajectory (green) using a discrete time LQG controller. Tracking this nominal trajectory incurs an obstacle collision probability of 0.4\%, which may be efficiently evaluated along with a tight associated confidence interval using variance-reduced Monte Carlo techniques (for this example in under 2 seconds to high confidence, a fraction of the 13 second trajectory duration).}
    \label{fig:plane_crashes}
\end{figure}

Specifically, the present work resides within the context of online robot trajectory planning under uncertainty subject to performance/safety constraints. This problem has been recognized recently 
as a critical component towards deploying autonomous robotic systems in unstructured environments 
\cite{ONR2012,Dahm2010,BergAbbeelEtAl2011}. Conceptually, to enable an autonomous system to plan its actions under uncertainty (e.g., with respect to environment characterization), one needs to design a strategy (i.e., closed-loop policy) for a decision maker \cite{LaValle2006}, a computationally expensive procedure in general \cite{JansonSchmerlingEtAl2015b}. Instead of optimizing over closed-loop policies that specify a next action for every possible history of measurements and control actions, a promising approach is to pose the decision-making problem as an optimization over a simpler class of \emph{nominal} policies, which are then evaluated via closed-loop predictions under the assumption that a local feedback control law strives to ensure nominal behavior \cite{JansonSchmerlingEtAl2015b,PatilBergEtAl2012,SunTorresEtAl2013,BergAbbeelEtAl2011}. This approach to planning is computationally much more tractable as it only involves optimization over open-loop action sequences, and represents a compromise between POMDP formulations involving minimization over the class of output-feedback control laws, and an open-loop formulation, in which the state is assumed to evolve in an open loop (i.e., it is conservatively assumed that over the future horizon no effort will be made to compensate for disturbances or new events).

Despite this simplification of the optimization domain, incorporating constraints into the problem poses another significant challenge. In particular, inspired by recent results within the domain of Model Predictive Control (MPC) \cite{OldewurtelJonesEtAl2008}, we consider \emph{chance-constrained} problem formulations where we seek the lowest cost nominal policies subject to a probabilistic bound on performance/safety constraint violation. In this context, a key component in any planning algorithm is quickly and reliably evaluating the probability of constraint violation (henceforth, we will focus on collision avoidance and refer to such a probability as \emph{collision probability} -- CP -- with the understanding that a constraint set on feasible robot states can also model performance or other safety requirements).  This problem is difficult as safety/performance constraints do not usually possess an additive structure, i.e., they can not be expressed as the expectation of a summation of a set of random variables. This is typical, for example, for chance constraints on obstacle avoidance over the length of a state trajectory, and a fortiori for more complex constraints (expressed, e.g., as logical formulas). 
The goal of this paper is to design an algorithm for the {\em the fast and reliable evaluation of collision probabilities}, taking particular consideration of non-linear state space dynamics perturbed by non-Gaussian noise, with full translational and rotational rigid-body collision checking (as opposed to point-robot-based projections of workspace obstacles into the state space).

\emph{Related Work}:
Evaluating a controller's chance of constraint satisfaction while guiding a robot along a reference trajectory has hitherto primarily been considered in an approximate fashion.
As discussed in \cite{JansonSchmerlingEtAl2015b}, most previous methods essentially rely on two approaches. In the first approach (see, e.g.,  \cite{AoudeLudersEtAl2013, KothariPostlethwaite2013, LudersKothariEtAl2010, LudersKaramanEtAl2013}), referred to as the ``additive approach," a trajectory CP is approximately evaluated by using Boole's inequality (i.e., $\prob{\cup_i A_i} \leq \sum_i \prob{A_i}$) and \emph{summing} pointwise CPs at a certain number of  waypoints along the reference trajectory. In contrast, in the second approach (see, e.g., \cite{LiuAng2014, ParkParkEtAl2016, SunTorresEtAl2013, BergAbbeelEtAl2011}), referred to as the ``multiplicative approach," a trajectory CP  is approximately evaluated by \emph{multiplying} the complement of pointwise CPs at a certain number of  waypoints along the reference trajectory. In a nutshell, the additive approach treats waypoint collisions as mutually exclusive, while the multiplicative approach treats them as independent. Since neither decoupling assumption holds in general, nor is any accounting made for continuous collision checking between waypoints (although \cite{ParkParkEtAl2016} and \cite{SunTorresEtAl2013} do consider the case of general collision geometries which in some cases may be amenable to such discrete collision checking), such heuristic approaches can be off by large multiples and hinder the computation of feasible trajectories \cite{JansonSchmerlingEtAl2015b}. Even worse, it is shown in \cite{JansonSchmerlingEtAl2015b} that such approaches are asymptotically tautological, i.e., as the number of waypoints approaches infinity, a trajectory CP is approximated with a number {\em greater than or equal to one}, contrary to what one might expect from a refinement procedure. (We note, however, that such approximations would still be very useful in those cases where the problem is unconstrained and the sole objective is to optimize safety/performance, as what is needed in such cases is the characterization of  {\em relative} CPs among trajectories. However, in a constrained setting, one needs an accurate, {\em absolute} measure of CP.) The work of \cite{PatilBergEtAl2012} introduced a first-order approximate correction to the independence assumption of the multiplicative approach that appears empirically to mitigate its aforementioned drawbacks, but can still be off by a considerable margin \cite{JansonSchmerlingEtAl2015b} and can both under-approximate or over-approximate the true CP.
 % (in contrast to the additive approach, which is always on the conservative side).

The limitations of the above approximation schemes motivate the approach in \cite{JansonSchmerlingEtAl2015b} to consider variance-reduced Monte Carlo (MC) to estimate trajectory CPs. In theory, the exact CP  can be computed up to arbitrary accuracy by simulating a large number of trajectories and counting the number that violate a constraint. When the CP has a low value (e.g., $\leq 1\%$), however, as is desired for most robotic applications, an enormous number of Monte Carlo samples may be required to achieve confidence in the estimate. The work in \cite{JansonSchmerlingEtAl2015b} demonstrates statistical variance-reduction techniques, including a method of importance sampling based on shifting the means of true noise distributions to encourage collision in simulation, that reduce the required number of samples to a few hundred per evaluation, amenable to real-time implementation.
However, these techniques are designed and validated only for linear time-invariant point-robot systems under the influence of Gaussian process and measurement noise. As both assumptions do not always hold in practice, one objective of this work is to extend variance-reduced MC methods to the case of non-linear dynamics and rigid-body collision models. Furthermore, the mixture IS algorithm proposed in \cite{JansonSchmerlingEtAl2015b} may also suffer in its ability to reduce variance as time discretization approaches zero (although it still provides an unbiased, asymptotically-exact estimator) due to correlated mixture terms corresponding to close potential collision points \cite{PanChittaEtAl2011}.

\emph{Statement of Contributions}:
The primary contribution of this paper is the proposal of an adaptive importance sampling framework (whereby the sampling distribution is refined
through the course of simulating sample trajectories) which enables accelerated Monte Carlo collision probability estimation
in general, non-linear problem settings with rigid-body collision checking that admits a high degree of correlated collision modes. This is a significant innovation (a) over the MC method in \cite{JansonSchmerlingEtAl2015b}, the naive application of which (i.e., linearizing dynamics and assuming a spherical robot) results in IS distributions with much poorer convergence characteristics, and (b) over approximation heuristics \cite{LiuAng2014, LudersKaramanEtAl2013, PatilBergEtAl2012, SunTorresEtAl2013, BergAbbeelEtAl2011} which are predicated on decoupling assumptions that are arguably over-simplistic even before the additional complexities of non-linear dynamics and non-convex robot bodies are added. Our approach is enabled (1) by an improved method for identifying likely collision points based on a Newton method from \cite{SunTorresEtAl2013}, and (2) by a more principled method for constructing IS distributions corresponding to certain collision modes, compared to the mean-shift method proposed in \cite{JansonSchmerlingEtAl2015b}, that enables optimization over more distributional parameters. Finally, we demonstrate the effectiveness of our adaptive IS framework 
on an 8-state, 3-input airplane model where it uses fewer MC samples than any alternative
to accurately evaluate, with an associated confidence margin constructed using a standard error estimate, a nominal policy with low associated CP ($0.4\%$). A single-threaded MC implementation achieves usable estimates with certified accuracy within 1-2 seconds ($< 500$ samples).
While in this paper we develop our approach assuming a discrete LQG controller equipped with an EKF (as considered in \cite{PatilBergEtAl2012,BergAbbeelEtAl2011}) applied to the linearized
dynamics and Gaussianized noise (first-and-second moment-matching) of a general problem setting as our closed-loop
feedback policy, we note that our adaptive framework applies equally well to other controllers; only the mathematical details of the distance metric and R\'enyi divergence objective in Section~\ref{sec:AIS}, detailing the selection of the initial IS distribution, will change. % mention static obstacles?
Given the successes of \cite{JansonSchmerlingEtAl2015b} in incorporating
MC collision probability estimation within a near-real-time safe motion planning algorithm, and especially in light of this MC method's
parallelization potential, %\cite{IchterSchmerlingEtAl2017}
we believe that the adaptive mixture importance sampling algorithm described in this paper represents
a tool worthy for consideration in a wide variety of robotic applications that plan with safety constraints in operation.

\emph{Organization:} The remainder of this paper is organized as follows: in Section~\ref{sec:background} we review background material on
importance sampling relevant to robot collision probability estimation including an adaptive technique, in Section~\ref{sec:problem} we define the problem dynamics, LQG controller model, and collision model. In Section~\ref{sec:AIS} we present how to tailor adaptive mixture IS to CP computation by constructing component distribution corresponding to likely collision modes. Section~\ref{sec:experiments} provides illustrative experiments with a rigid-body non-linear airplane model for the algorithms detailed in the previous sections, and Section~\ref{sec:conc} contains conclusions and discussion on possible directions of future research.

\section{Background Material} \label{sec:background}
In this section we review the basics of importance sampling, a technique for reducing the variance of a Monte Carlo
estimator, as well as an adaptive variant for mixture importance sampling that frames the selection of mixture weights as a convex optimization problem. Our discussion is tailored to the problem of computing trajectory CPs, where the high-dimensional nature of the noise (a joint distribution spanning the entire length of a trajectory) increases the challenge of selecting a good distribution.

\subsection{Importance Sampling} \label{subsec:IS}
Consider a random variable $\X \in \R^n$ distributed according to the probability density function (pdf) $P: \R^n \rightarrow \R_{\geq 0}$.
The expectation $\E{f(\X)} = \int_{R^n} f(x)P(x)\dx$ of a function $f(\X)$ may be rewritten as
\begin{equation}\label{eq:ISrewrite}
\begin{aligned}
\EP{f(\X)} &= \int_{R^n} f(x)P(x)\dx \\
           &= \int_{R^n} \left(f(x)\frac{P(x)}{Q(x)}\right)Q(x)\dx \\
           &= \EQ{f(\X)\frac{P(\X)}{Q(\X)}},
\end{aligned}
\end{equation}
where $Q$ is an alternative pdf satisfying $Q(x) > 0$ for all $x \in \R^n$ where $f(x)P(x) \neq 0$, and $\mathbb{E}^P$ and 
$\mathbb{E}^Q$ denote expectations computed under the distributions $\X \sim P$ and $\X \sim Q$, respectively. The quantity
$w(x) = P(x)/Q(x)$ is referred to as the likelihood ratio of the pdfs at $x$.

In the context of Monte Carlo estimation, the above computation implies that given independent and identically distributed (i.i.d.) samples $\{\X^{(i)}\}_{i=1}^m$
drawn from $Q$, the expectation $p := \E{f(\X)}$ may be estimated by
\begin{equation}\label{eq:ISp}
\begin{aligned}
\hat p^Q &:= \frac{1}{m}\sum_{i=1}^m f(\X^{(i)})w(\X^{(i)}).
\end{aligned}
\end{equation}
The variance of the estimator $\hat p^Q$ may be estimated by
\begin{equation}\label{eq:ISV}
\hat V^Q := \frac{1}{m^2}\sum_{i=1}^m \left(f(\X^{(i)})w(\X^{(i)}) - \hat p^Q\right)^2.
\end{equation}
In this paper, $\X$ will denote a vector of noise samples that act upon a robot at a sequence of discrete time increments,
and $f(\X)$ will denote the indicator function that the robot, guided by a trajectory-tracking controller, collides into
the obstacle set given the trajectory noise sample $\X$. The robot's collision probability is $p = \E{f(\X)}$.

We see from \eqref{eq:ISrewrite} that $\hat p^Q$ defined in \eqref{eq:ISp} is an unbiased estimator of the CP $p$, that
is, as long as the support of $Q$ contains the support of $P$, the estimator $\hat p^Q$ converges in probability to $p$ as
$m \rightarrow \infty$ regardless of the choice of $Q$. Thus the selection criterion for the IS pdf $Q$ is to minimize
the variance of $\hat p^Q$. This has the practical effect of reducing the number of samples required before the estimator
yields a useable result, which, as argued in \cite{JansonSchmerlingEtAl2015b}, is critical for real-time accurate CP estimation for
robotic applications with a very low collision chance constraint. Motivated by the sum in \eqref{eq:ISp}, we call $\VarQ{f(\X)\frac{P(\X)}{Q(\X)}}$ the per-sample variance contributed by each i.i.d. sample $\X^{(i)} \sim Q$. Choosing $Q$ to minimize the per-sample variance is equivalent \cite{Ryu2016} to minimizing the R\'enyi divergence
\begin{equation}\label{eq:renyi}
D_2\left(\pi^* \| Q\right) := \log \int_{R^n} \frac{\pi^*(x)^2}{Q(x)}\dx
\end{equation}
where $\pi^*(x) := |f(x)|P(x)/\int_{R^n} |f(x)|P(x)\dx$ is itself the minimizer, provided the optimization is unconstrained. 

Constructing and sampling from $\pi^*$ is usually not possible in practice (for positive $f$, the normalization factor $\EP{f(\X)}$ is precisely
the quantity we wish to estimate), but its form yields some insights. In the case that $f$ is an indicator function,
$\pi^*$ has support on only the ``important'' parts of $P$ where $f(x) = 1$; for our purposes, to sample from $\pi^*$ is
to sample only noise trajectories that lead to collision and weight them in the computation of $\hat p^{\pi^*}$ by their
relative likelihood. This motivates the search for IS distributions $Q$ that artificially inflate the occurrence of the rare
event $f(\X) = 1$, but this should be accomplished while maintaining relative probability according to $P$ lest the
likelihood ratio $P/Q$ be very large for some likely realization of the event, corresponding to a large value in the
variance integral~\eqref{eq:renyi}. Mathematically, we attempt to minimize $D_2\left(\pi^* \| Q_\theta\right)$ over a family of distributions $\mathcal{Q} = \{Q_\theta \mid \theta \in \Theta\}$, described by a finite vector of parameters $\theta$, capable of capturing this aim.
\subsection{Mixture Importance Sampling}\label{subsec:mixtureIS}
As recognized in \cite{JansonSchmerlingEtAl2015b}, there are typically
multiple ways in which a noise-perturbed robot trajectory can collide with its surroundings. Although the noise pdf $P$
is usually unimodal (corresponding to a robot centered on the nominal trajectory), an effective IS distribution may be multimodal (corresponding to the many ways the robot can drift into obstacles). This motivates the use of
mixture IS distributions with pdfs of the form
\begin{equation}\label{eq:mixtureIS}
Q_{(\alpha, \eta)}(x) = \sum_{d=1}^D \alpha_d q_d(x; \eta_d),
\end{equation}
parameterized by $\theta = (\alpha, \eta)$; the $\alpha_d$ are nonnegative mixture weights such that $\sum_{d=1}^D \alpha_d = 1$ and the $\eta_d$ are internal
parameters of the component densities $q_d$. A special case of mixture IS relevant to robotic applications is defensive importance sampling, where the nominal distribution $P$ is selected as one of the component distributions. If the pdfs $q_d$ are selected as noise likely to lead to certain collision scenarios, including $P$ serves as a catch-all to ensure that no unforeseen collision mode, e.g., due to the complex evolution of uncertainty distributions through non-linear dynamics, is left out.

\subsection{Adaptive Mixture Importance Sampling} \label{subsec:AIS}
For robot collision probability estimation, optimizing the per-sample variance over a full family of distributions~\eqref{eq:mixtureIS} is in general computationally intractable, especially if the component trajectory noise parameterization $\eta_d$ is high-dimensional. Thus we consider instead the problem of selecting the weights $\alpha_d$ for fixed components $q_d$. As shown in \cite{Ryu2016}, the objective $D_2\left(\pi^* \| Q_\alpha\right)$ is convex with respect to $\alpha$, and is therefore amenable to online stochastic optimization methods with convergence guarantees. We reproduce in Algorithm~\ref{alg:mixtureIS} a stochastic mirror descent procedure from \cite{Ryu2016}, designed for the simultaneous adaptation of mixture distribution weights alongside IS estimation. Algorithm~\ref{alg:mixtureIS} performs stochastic gradient descent on a set of mirrored variables $\tilde\alpha$ in order to enforce the probability constraints $\sum_{d=1}^D \alpha_d = 1,\ \alpha_d \geq 0\ \forall d$. We present here the self-normalized versions of the final estimators, where in computing $\hat p$ we normalize by the sum of the sampled likelihood ratios. This is an alternative to normalizing, as in~\eqref{eq:ISp}, by the reciprocal of the sample count $1/m$. Self-normalized importance sampling yields an asymptotically unbiased estimator, and in practice may further reduce variance when applying importance sampling techniques. We note that normalizing by the sample count $m = k\ell$ would only change the form of expressions in Alg.~\ref{alg:mixtureIS}, Line~\ref{line:estimators}.

\begin{algorithm}[t]
\caption{Adaptive Mixture IS (Section 3.4.4, \cite{Ryu2016})}
\label{alg:mixtureIS}
\begin{algorithmic}[1]
\REQUIRE Component densities $q_1(x),\dots,q_D(x)$, initial mixture weights $\alpha^1$, step size parameter $C$, batch size $k$, number of iterations $\ell$
\STATE Set mirrored weights: $\widetilde{\alpha^{1}} = \log(\alpha^{1})$
\FOR{$i = 1:\ell$}
\STATE Sample $\{\X_{i,j}\}_{j=1}^k$ from current IS distribution $Q_{\alpha^i}$
\STATE Compute gradient:
\[
g_i = -\frac{1}{k}\sum_{j=1}^k \frac{\left(f(\X_{i,j})\frac{P(\X_{i,j})}{Q_{\alpha^i}(\X_{i,j})}\right)^2}{Q_{\alpha^i}(\X_{i,j})} \begin{bmatrix}q_1(\X_{i,j})\\ \vdots \\q_D(\X_{i,j})\end{bmatrix}
\]
\STATE Update mirrored weights: $\widetilde{\alpha^{i+1}} = \widetilde{\alpha^{i}} - (C/\sqrt{i})g_i$
\STATE Set new mixture weights: $\alpha^{i+1} \propto \exp\left(\widetilde{\alpha^{i+1}}\right)$
\ENDFOR
\RETURN \label{line:estimators} Estimator, estimated variance of estimator \cite{Owen2013}:
\begin{align*}
\hat p^{\text{AIS}} &= \frac{\sum_{i=1}^\ell \sum_{j=1}^k f(\X_{i,j})\frac{P(\X_{i,j})}{Q_{\alpha^i}(\X_{i,j})}}
                            {\sum_{i=1}^\ell \sum_{j=1}^k \frac{P(\X_{i,j})}{Q_{\alpha^i}(\X_{i,j})}}\\
\hat V^{\text{AIS}} &= \frac{1}{k\ell}\frac{\sum_{i=1}^\ell \sum_{j=1}^k \left(\frac{P(\X_{i,j})}{Q_{\alpha^i}(\X_{i,j})}(f(\X_{i,j}) - \hat p^{\text{AIS}})\right)^2}
                                           {\left(\sum_{i=1}^\ell \sum_{j=1}^k \frac{P(\X_{i,j})}{Q_{\alpha^i}(\X_{i,j})}\right)^2}
\end{align*}
\end{algorithmic}
\end{algorithm}

\section{Problem Formulation} \label{sec:problem}

\subsection{State Space Dynamics}
Let the state space dynamics of a robot be given by
\begin{equation}\label{eq:truedynamics}
\x_{t} = \f(\x_{t-1}, \u_{t-1} + \v^u_t) + \v^x_t, \quad \z_t = \h(\x_t) + \w_t,
\end{equation}
where $\x_t \in \R^{d_x}$ is the state, $\u_t \in \R^{d_u}$ is the control input, $\z_t \in \R^{d_z}$ is the
measurement, $\v_t = [\v^u_t; \v^x_t]$ is the process noise, and $\w_t \sim \W_t$ is the measurement noise at time $t$. The process noise comprises an explicit control uncertainty $\v^u_t \sim \V^u_t$ in addition to a propagation uncertainty $\v^x_t \sim \V^x_t$; we note that in the case of a non-linear transition function $\f$ even Gaussian $\V^u_t$ may generate non-Gaussian state uncertainty at each time step.
We restrict our attention to independent noise distributions $\V^u_t$, $\V^x_t$, and $\W_t$ with zero mean and finite second
moments, noting that cases of
colored noise may be addressed through state augmentation. Let $V^u_t$, $V^x_t$, and $W_t$ denote the covariance matrices of $\V^u_t$, $\V^x_t$, and $\W_t$, respectively.
We are interested in
estimating the collision probability that arises from tracking a nominal path $\mathcal{P}^* = \left\{\x_0^*, \u_0^*,
\x_1^*, \u_1^*, \dots, \x_T^*\right\}$ where $\x_{t}^* = \f(\x^*_{t-1}, \u^*_{t-1})$ for $t = 1, \dots, T$. The true initial
state $\x_0$ satisfies $\x_0 = \x_0^* + \pp_0$, where the initial state uncertainty $\pp_0 \sim \mathbf{P}_0$
is drawn from a distribution with zero mean and covariance matrix $P_0$, and the controller uses an understanding of the dynamics and information from
observations $\z_1, \cdots, \z_{T}$ to choose actions with feedback $\u_0, \dots, \u_{T-1}$ from which the true state $\x_t$ evolves
according to \eqref{eq:truedynamics}.

Similar to \cite{PatilBergEtAl2012} and \cite{BergAbbeelEtAl2011} we consider an LQR state-feedback controller equipped with an
extended Kalman Filter for state estimation (together, LQG control) to control the deviation of a robot from a nominal
trajectory. With deviation variables from $\mathcal{P}^*$ defined as $\xb_t = \x_t - \x_t^*$, $\ub_t = \u_t - \u_t^*$, and
$\zb_t = \z_t - \h(\x_t^*)$, we linearize \eqref{eq:truedynamics} according to:
\begin{equation}\label{eq:lindynamics}
\begin{aligned}
\xb_{t} &= A_{t} \xb_{t-1} + B_{t} \ub_{t-1} + (B_{t}\v^u_t + \v^x_t)\\
         &\qquad+ O(\|\xb_{t-1}\|^2 + \|\ub_{t-1}\|^2), \\
\zb_t &= H_t \xb_t + \w_t + O(\|\xb_t\|^2),
\end{aligned}
\end{equation}
where $A_t = \frac{\partial \f}{\partial \x}(\x_{t-1}^*, \u_{t-1}^*)$,
$B_t = \frac{\partial \f}{\partial \u}(\x_{t-1}^*, \u_{t-1}^*)$,
and $H_t = \frac{\partial \h}{\partial \x}(\x_{t}^*)$. The LQG controller maintains an estimate $\xh_t$ of the true state
deviation $\xb_t$ using an extended Kalman filter
\[
\xh_{t} = K_{t}\zb_{t} + (I - K_{t} H_{t}) (A_{t} \xh_{t-1} + B_{t} \ub_{t-1}),
\]
where $K_t$ is the Kalman gain matrix at time $t$ corresponding to the linearized system \eqref{eq:lindynamics} with
noise covariance matrices $P_0$, $V_t = \begin{bmatrix}B_t V^u_t B_t^T & 0 \\ 0 & V^x_t\end{bmatrix}$, and $W_t$ \cite{Haykin2001}. Then at each time $t = 0, \dots, T-1$ the LQG controller applies
the input $\u_t = \u_t^* + \ub_t$ with
\[
\ub_t = \u_t^* + L_{t+1} \xh_t,
\]
where $L_t$ is the finite time horizon LQR feedback gain matrix corresponding to \eqref{eq:lindynamics} with appropriately
chosen state regulation and control effort penalties for the robotic application.

We emphasize that when simulating trajectories in this paper, we compute the exact state evolution according to the true dynamics
\eqref{eq:truedynamics}. However, we note here that with $\y_t = \begin{bmatrix}\xb_t ; \xh_t\end{bmatrix}$ we may write down
an approximate system with linearized dynamics and Gaussian noise (moment-matched to the true noise up to second order):
\begin{equation}\label{eq:linapprox}
\y_{t} = F_t \y_{t-1} + G_t \mathbf{r}_t, \quad \mathbf{r}_t \sim \mathcal{N}(\0, R_t),
\end{equation}
for appropriate choices of matrices $F_t, G_t$ and $R_t$. See \cite{BergAbbeelEtAl2011} for a full derivation. In addition to
evolving approximate state trajectories, this system may be used to evolve approximate pointwise uncertainty distributions of $\xb_t$ parameterized as
multivariate Gaussians, as in \cite{JansonSchmerlingEtAl2015b,PatilBergEtAl2012,BergAbbeelEtAl2011}. Let $\Sigma_t$ denote the a priori covariance of $\xb_t$ thus derived.

\subsection{Configuration Space and Workspace Representations}\label{subsec:csws}
In this work we consider rigid-body robots whose configuration $\q_t = \q(\x_t) \in SE(3)$, consisting of a 3D rotation and translation, is a deterministic function of the state. We represent both the robot, configured at $\q$, and the static obstacle set in the workspace as unions of convex components $\mathcal{R}(\q) = \bigcup_{i=1}^r \mathcal{R}_i(\q) \subset \R^3$ and $\mathcal{E} = \bigcup_{j=1}^e \mathcal{E}_j \subset \R^3$, respectively. Then the state space obstacle set is $\Xobs = \{\x \in \R^{d_x} \mid \mathcal{R}(\q(\x)) \bigcap \mathcal{E} \neq \varnothing\}$. For a given robot state $\x$ and associated configuration $\q$, we assume access to a distance function $d_{i,j}(\q)$ measuring the Euclidean separation between $\mathcal{R}_i(\q)$ and $\mathcal{E}_j$. In the case that $\mathcal{R}_i(\q)$ intersects $\mathcal{E}_j$, $d_{i,j}(\q)$ returns a negative value corresponding to the maximum extent of penetration. We also assume access to the distance gradient $\partial d_{i,j}(\q)/\partial \q$, either analytically or through finite differencing. We may also compute $\partial d_{i,j}(\x)/\partial \x = \frac{\partial\q}{\partial\x}\frac{\partial d_{i,j}}{\partial\q}$.

\subsection{Problem Statement}
The problem we wish to solve in this paper is to devise an accurate, computationally-efficient algorithm equipped with an error estimate to estimate the collision probability
\[
\prob{\overline{\x_0,\dots,\x_T} \cap \Xobs \neq \varnothing}
\]
where $\overline{\x_0,\dots,\x_T}$ denotes a continuous interpolation between discrete states, and the state trajectory $\x_t$ is controlled via the control law $\u_t = \u_t^* + \ub_t$. As discussed in the introduction, the primary motivation of this problem is to enable belief space planning with closed-loop predictions for general non-linear problems with possibly non-Gaussian noise models. 

A few comments are in order. First, in this paper we are not proposing a new planning algorithm, rather an algorithm that addresses one of the key bottlenecks for planning under uncertainty. Second, the method proposed here can be used in combination with a variety of planning frameworks, e.g., to certify the final plans output by the heuristic evaluation of many RRT plans \cite{SunTorresEtAl2013,BergAbbeelEtAl2011} or by bounded uncertainty roadmaps \cite{GuibasHsuEtAl2008}, or within the planning loop of meta-algorithms as in \cite{JansonSchmerlingEtAl2015b}. Third, although for clarity Subsection \ref{subsec:csws} assumes an $SE(3)$ configuration space and $\R^3$ workspace, the methods in this paper may be readily generalized to other rigid-body robots, e.g., manipulators \cite{SunTorresEtAl2013}. Finally, in stark contrast with alternative methods (with the exception of the MC method \cite{JansonSchmerlingEtAl2015b}), we provide a computable error estimate that can be used as a certificate of accuracy for the trajectory's estimated CP.

\section{Adaptive Importance Sampling for \\Collision Probability Estimation} \label{sec:AIS}

\begin{algorithm}[t]
\caption{Close Pairwise $\Xobs$ Point (adapted from \cite{SunTorresEtAl2013})}
\label{alg:cop}
\begin{algorithmic}[1]
\REQUIRE Nominal mean $\x^* \in \R^{d_x}$, covariance matrix $\Sigma \in \R^{d_x \times d_x}$, workspace distance function $d_{i,j}(\q(\x))$ between robot/environment pair of convex components $\mathcal{R}_i(\q)$ and $\mathcal{E}_j$, linesearch parameter $\gamma$, tolerance $\epsilon > 0$
\STATE Set $\x_0 = \x^*, k = 0$
\REPEAT
\STATE Newton step (derivatives evaluated at $\x_k, \q(\x_k)$):
{\small
\[
\mkern-40mu\x_{k+1} = \x_k - d_{i,j}(\x_k)\Sigma\frac{\partial d_{i,j}}{\partial\x}(\x_k)\bigg/
            \left(\frac{\partial d_{i,j}}{\partial\x}(\x_k)^T\Sigma\frac{\partial d_{i,j}}{\partial\x}(\x_k)\right)
\]
}
\STATE $k = k + 1$
\UNTIL{$\|\x_k - \x_{k-1}\| < \epsilon$}
\STATE $m_k = (\x_k - \x^*)^T\Sigma^{-1}(\x_k - \x^*)$
\REPEAT
\STATE With $\g = \frac{\partial d_{i,j}}{\partial\x}(\x_k)$, compute search direction:
\[
\s = \Sigma^{-1}\g - \left(\frac{\g^T\Sigma^{-1}(\x_k - \x^*)}{\g^T\g}\right)\g
\]
\STATE $\alpha = 1$
\REPEAT
\STATE \label{line:proj}$\x_{k+1} = \mathrm{project}_{i,j}\left(\x_k - \frac{\alpha\gamma m_k \s}{\s^T \Sigma^{-1}(\x_k - \x^*)}\right)$
\STATE $m_{k+1} = (\x_{k+1} - \x^*)^T\Sigma^{-1}(\x_{k+1} - \x^*)$
\STATE $\alpha = \alpha/2$
\UNTIL{$m_{k+1} \leq m_{k}$ or $\|\x_{k+1} - \x_{k}\| < \epsilon$}
\STATE $k = k + 1$
\UNTIL{$m_k \geq m_{k-1}$ or $\|\x_k - \x_{k-1}\| < \epsilon$}
\RETURN $\x_k$ with minimum corresponding $m_k$
\end{algorithmic}
\end{algorithm}

In this section we present an algorithm for the accurate, computationally-efficient estimation of a trajectory's tracking CP under non-linear dynamics, full rigid-body collision checking, and non-Gaussian noise models.  
Our approach is to select importance sampling distributions for CP estimation as a mixture of reparameterized copies of the actual process
and measurement noise, corresponding to different modes of failure, similar to~\cite{JansonSchmerlingEtAl2015b}. In particular we employ assumptions of linearity and Gaussianity to compute collision-causing adjustments to the noise means and covariances. A key strength of the MC approach is that although these assumptions do not necessarily reflect the true system dynamics, the IS distributions they suggest are still useful in the context of Algorithm~\ref{alg:mixtureIS} (an asymptotically unbiased estimator in any case) to perform variance-reduced CP estimation.

In the notation of
Section~\ref{sec:background},
\begin{align*}
\X &= \left(\pp_0, \v^u_1, \v^x_1, \w_1, \dots, \v^u_{T}, \v^x_T, \w_{T}\right),\\
P(x) &= \mathbf{P}_0(p_0) \cdot \V^u_1(v^u_1) \cdot \ldots \cdot \W_{T}(w_{T}),
\end{align*}
(where we have assumed the independence of the noise distributions in the construction of this joint pdf) and $f(\X)$ is
the event that the noise random variable $\X$ gives rise to a colliding trajectory under LQG control. We consider IS
distributions of the form
\begin{align*}
Q_{(\alpha, \eta)}(x) &= \sum_{d=1}^D \alpha_d q_d(x; \eta_d)\\
q_d(x; \eta_d) &= \mathbf{P}_0(p_0; \eta_d) \cdot \V^u_1(v^u_1; \eta_d) \cdot \ldots \cdot \W_{T}(w_{T}; \eta_d)
\end{align*}
where $\eta_d$ encodes all of the parameters required to specify the process and measurement noise distributions. We will consider IS component distributions $q_d$ that differ from the actual noise $P$ only in mean and covariance (e.g., produced by affine transformation), and thus in the following discussion for simplicity we consider $\eta_d$ to consist of $(3T+1)$ mean vectors and covariance matrices: one pair for
the initial state uncertainty and for each process/measurement noise distribution at each time step $t = 1, \dots, T$. We pack these means and covariances into $(T+1)$ Gaussians acting on the linearized dynamics~\eqref{eq:linapprox}
and write (omitting the index $d$), $\eta = \left((\mub_{0}, S_{0}), \dots, (\mub_{T}, S_{T})\right)$, where for time step $t$, $(\mub^{v_u}_t, \Sigma^{v_u}_t)$, $(\mub^{v_x}_t, \Sigma^{v_x}_t)$, $(\mub^{w}_t, \Sigma^{w}_t)$ correspond to
\[
\mub_{t} = \begin{bmatrix} B_t \mub^{v_u}_t + \mub^{v_x}_t \\ K_t \mub^{w}_t \end{bmatrix}, S_{t} = \begin{bmatrix}B_t \Sigma^{v_u}_t B_t^T + \Sigma^{v_x}_t & 0 \\ 0 & K_t \Sigma^{w}_t K_t^T\end{bmatrix}.
\]
We stress that this linearization is a step that we take only in order to derive expressions for $\eta$ below,
but which cannot be done when MC simulating the full non-linear dynamics~\eqref{eq:truedynamics}. We will also refer to $\eta = (\mub, S)$ as the parameters of the joint Gaussian over all trajectory time steps.

\subsection{Computing Likely Collision Modes}\label{subsec:closepoints}

We choose each component distribution $q_d$ to represent a likely tracking collision mode (see Figure~\ref{fig:plane_crashes}). In particular, at each time step $t$ we consider distributions $q$ that result in an expected collision (under the linearized dynamics~\eqref{eq:linapprox}) at an obstacle point $\x_\text{obs} \in \Xobs$ close to $\x^*_t$, that is
\[
\mathbb{E}^{q}\left[\xb_t\right] \approx \begin{bmatrix}I & 0\end{bmatrix} M_t \mub = \x_\text{obs} - \x_t^*
\]
where the matrix $M_t$, which describes the evolution of the mean of $\y_t$ according to~\eqref{eq:linapprox}, satisfies for all $\mub$:
\[
M_t \mub = \sum_{s = 0}^t \left(\prod_{r=0}^{t-s-1}F_t\right)G_t \mub_t.
\]
To ensure that our IS distribution represents the most likely collision modes, we define closeness to be 
measured by the Mahalanobis distance $\sqrt{(\x^*_t - \x_\text{obs})^T\Sigma_t^{-1}(\x^*_t - \x_\text{obs})}$, a measure of likelihood under the unmodified noise $P$. The problem of computing close obstacle points under this metric for a robot consisting of convex rigid-body components was previously considered in \cite{SunTorresEtAl2013}, which, for a pair $(\mathcal{R}_i, \mathcal{E}_j)$, proposes a Newton method for identifying a close point $\x_\text{obs}$ satisfying $d_{i,j}(\q(\x_\text{obs})) = 0$. We apply that method in Algorithm~\ref{alg:cop} to first identify a feasible point satisfying the zero-distance constraint. Our addition in Algorithm~\ref{alg:cop} is to then follow this initialization with a non-linear constrained minimization phase to find a local optimum in Mahalanobis distance. This minimization phase employs a linesearch with a constraint projection subroutine ($\text{project}_{i,j}$ Alg.~\ref{alg:cop}, Line~\ref{line:proj}) to ensure $d_{i,j}(\q(\x_\text{obs})) = 0$; we use the same Newton method to implement that projection as well. In our experiments we find that local optimization can reduce $\x_\text{obs}$ Mahalanobis distance by $\sim5\%$.

\begin{figure}[t]
\centering
    \subfigure[]{\label{fig:copall} \includegraphics[width=0.23\textwidth]{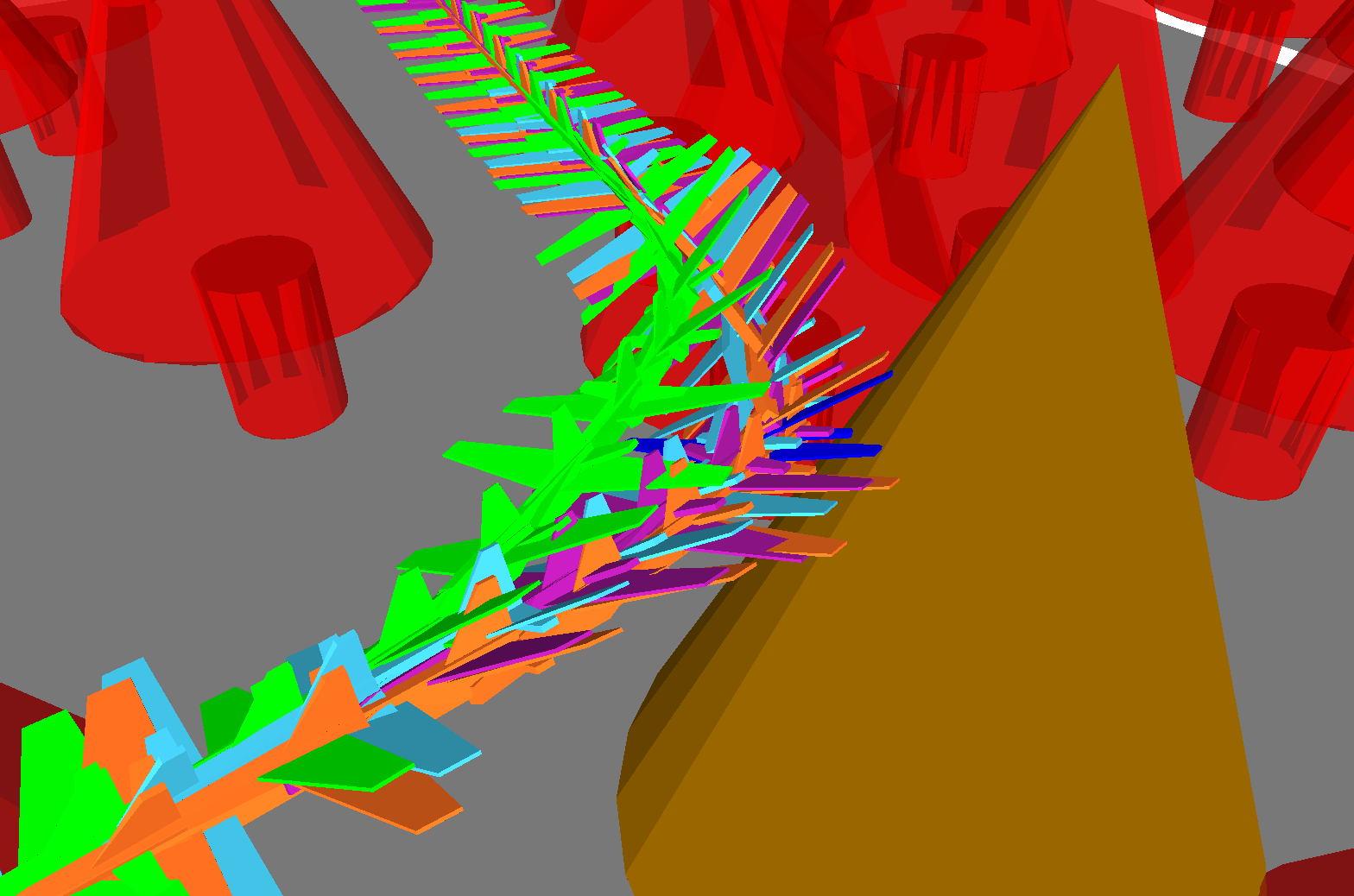}}
    \subfigure[]{\label{fig:cop1} \includegraphics[width=0.23\textwidth]{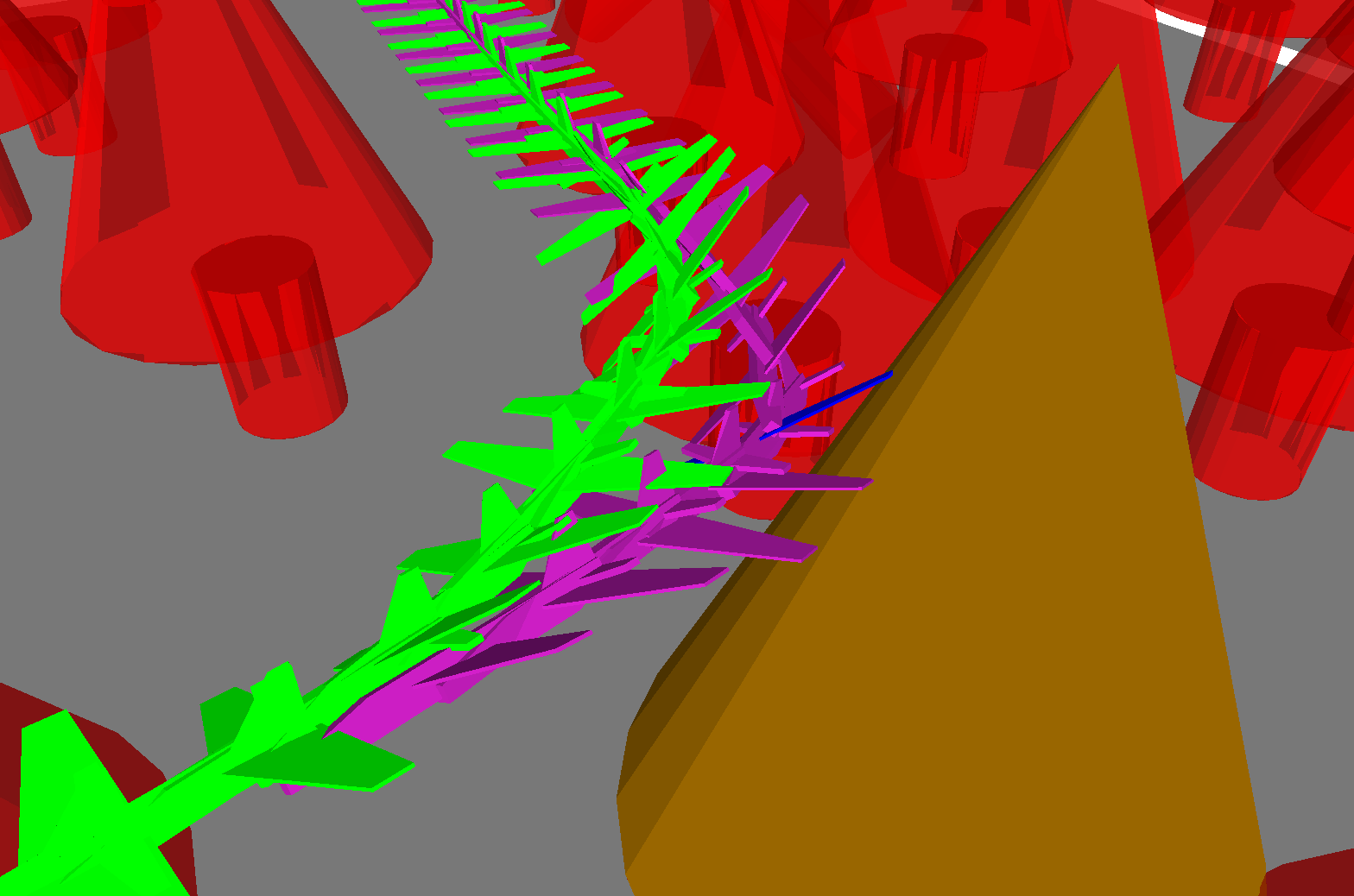}}
    \subfigure[]{\label{fig:cop2} \includegraphics[width=0.23\textwidth]{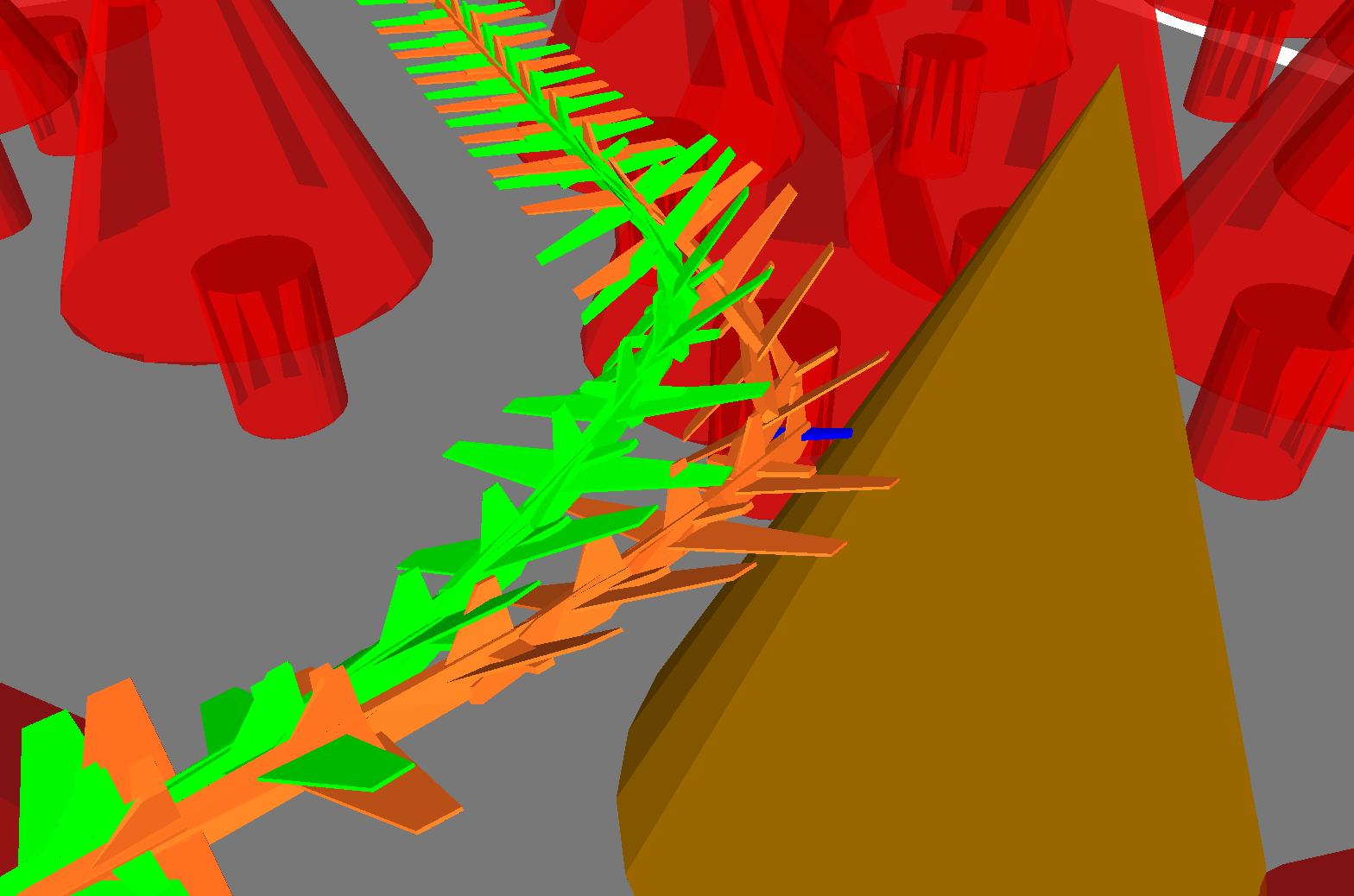}}
    \subfigure[]{\label{fig:cop3} \includegraphics[width=0.23\textwidth]{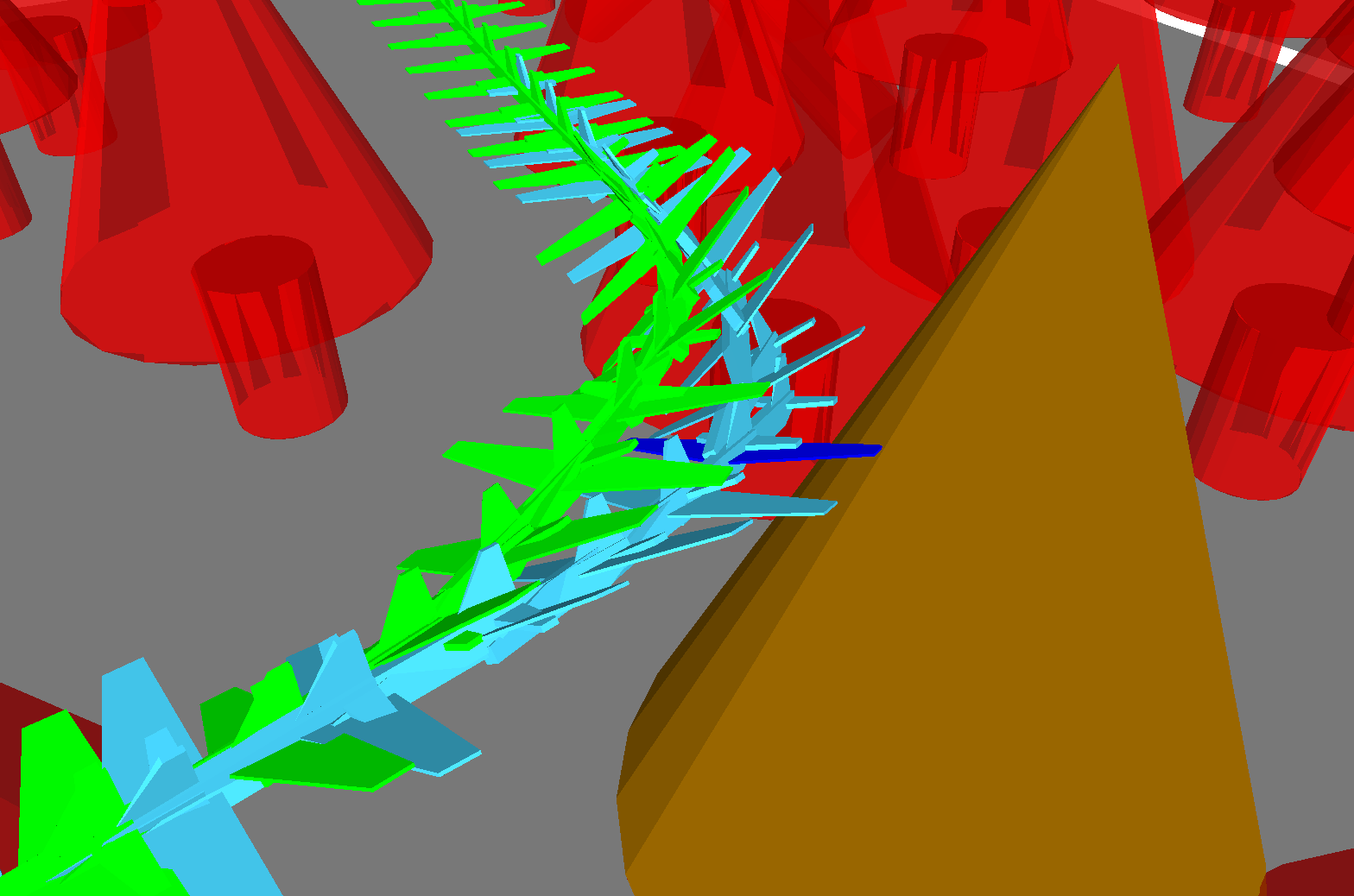}}
        \caption{The three most likely collision points, measured by Mahalanobis distance, for the nominal trajectory displayed in Figure~\ref{fig:plane_crashes}. Figure \ref{fig:copall} displays the high degree of correlation between the likely collisions \ref{fig:cop1} wing strike at $t = 82$, \ref{fig:cop2} stabilizer strike at $t = 83$, and \ref{fig:cop3} wing strike at $t = 83$. Convex robot components in collision are highlighted in blue.}
    \label{fig:cops}
\end{figure}

\subsection{Noise Distribution Corresponding to a Collision Mode}
Each close point $\x_\text{obs}$, corresponding to a tuple $(\x_t, \mathcal{R}_i, \mathcal{E}_j)$, may be viewed as a local proxy for the optimal IS distribution $\pi^*$. Recall from Section~\ref{sec:background} that we wish to compute IS distributions $q(x; \eta)$ such that $\eta = \argmin_\eta D_2\left(\pi^* \| q(;\eta)\right)$. We consider the optimization problem
\begin{equation}\label{eqn:ISoptgen}
\begin{aligned}
\argmin_{\eta}\ \ & D_2(P \| q(;\eta)) \\
\text{s.t.}\ \ & \mathbb{E}^{q}\left[\xb_t\right] = \x_\text{obs} - \x_t^*
\end{aligned}
\end{equation}
which, allowing for linearization and Gaussian moment-matching, may be written out explicitly as
\begin{equation}\label{eqn:ISoptgauss}
\begin{aligned}
\argmin_{\mub, S}\ \ & \mub^T (2S-R)^{-1} \mub - \frac{1}{2} \log\left(\frac{|2S - R||R|}{|S|^2}\right) \\
\text{s.t.}\ \ & \begin{bmatrix}I & 0\end{bmatrix} M_t \mub = \x_\text{obs} - \x_t^*
\end{aligned}
\end{equation}
where we have denoted the joint distribution of the linearization of the nominal noise $P$ (see~\eqref{eq:linapprox}) as $\mathcal{N}(\0, R)$. The IS method described in \cite{JansonSchmerlingEtAl2015b} may be regarded as a special case of this optimization with $S = R$ held fixed; then~\eqref{eqn:ISoptgauss} becomes a linearly constrained least squares problem where the objective $\mub^T R^{-1} \mub$ has the interpretation that $\eta = (\mub, S = R)$ is the true noise $P$ with mean equal to the most likely sample from $P$ that pushes $\x_t$ into $\x_\text{obs}$ (in expectation). We initialize optimization with this choice of $\eta$; to optimize over the covariance matrix $S$ as well, yielding greater variance reduction, we apply block coordinate descent: alternating optimization over $\mub$ with $S$ fixed (a linearly constrained least squares problem) and optimization over $S$ with $\mub$ fixed (we apply a simple first-order linesearch on the objective, as the constraint is automatically satisfied with $\mub$ fixed).

\subsection{Selecting Mixture Components}

The close point computation in \ref{subsec:closepoints} may theoretically give rise to $T r e$ different component distributions (one for each selection of time step, convex robot component, and convex obstacle), but we prune the mixture distribution $Q_{(\alpha,\eta)}$ to contain only components corresponding to each of the top $D-1$ most likely collisions computed over tuples $(\x_t, \mathcal{R}_i, \mathcal{E}_j)$, ordered by Mahalanobis distance (we set the final term $q_D = P$ to enable defensive importance sampling). The initial mixture weights $\alpha^1$ may be set uniformly, or in proportion to the probability that a state sampled from a Gaussian estimate of the marginal deviation distribution at the relevant trajectory time step crosses a half-space associated with each close obstacle point \cite{PatilBergEtAl2012,SunTorresEtAl2013}. We note that the value of $D$ may be chosen to correspond to a cutoff on these pointwise collision probability estimates as opposed to being a fixed algorithm parameter. 
Figure~\ref{fig:cops} illustrates why adaptive mixture IS, Algorithm~\ref{alg:mixtureIS}, is a necessary addition for refining these weights online to achieve the highest degree of variance reduction. For this plane trajectory, there is a high degree of correlation between similar collision events at successive time steps, or at the same time step between collisions involving different robot components. Unlike any heuristic method to address these correlations, which also grow worse with finer time discretization, stochastically solving the convex optimization problem of mixture weight selection is guaranteed to converge to the optimal $\alpha^*$ as the sample size $m \rightarrow \infty$.

\section{Numerical Experiments} \label{sec:experiments}
\subsection{Dynamics and Noise Model}
In this paper we consider discrete-time nominal dynamics for an airplane integrated from a simple continuous-time model \cite{BeardMcLain2012} propagated under zero-order hold control inputs with a time step $\Delta t$. The continuous-time model is:
\begin{equation}
\begin{aligned}
\dot\x = \begin{bmatrix}
\dot x\\
\dot y\\
\dot z\\
\dot v\\
\dot \psi\\
\dot \gamma\\
\dot \phi\\
\dot \alpha
\end{bmatrix} = 
\begin{bmatrix}
v \cos(\psi) \cos(\gamma)\\
v \sin(\psi) \cos(\gamma)\\
v \sin(\gamma)\\
u_a - F_\text{drag}(v,\alpha)/m - g \sin(\gamma)\\
-F_\text{lift}(v,\alpha) \sin(\phi)/(m v\cos(\gamma))\\
F_\text{lift}(v,\alpha) \cos(\phi)/(m v) - g \cos(\gamma)/v\\
u_{\dot\phi}\\
u_{\dot\alpha}
\end{bmatrix},
\end{aligned}
\end{equation}
where $x,y,z$ are position in a global frame, $v$ is airspeed (we assume zero wind, aside from isotropic gusts represented by process noise), $\psi$ is the course angle, $\gamma$ is the flight path angle, $\phi$ is the roll angle, and $\alpha$ is the angle of attack.
The control inputs $\u = (u_a, u_{\dot\phi}, u_{\dot\alpha})$ are longitudinal acceleration (due to engine thrust), roll rate, and pitch rate respectively. We assume a flat-plate airfoil model so that $F_\text{lift} = \pi\rho A v^2\alpha$ and $F_\text{drag} = \rho A v^2 (C_{D_0} + 4\pi^2K\alpha^2)$ \cite{BeardMcLain2012} where gravity $g$, air density $\rho$, wing area $A$, plane mass $m$, drag coefficient $C_{D_0}$, and induced drag factor $K$ are all constants.
Using the Euler ZYX (yaw $\psi$, pitch $\theta$, roll $\phi$) rotation angle convention, the mapping from state space to configuration space is given by $\q(\x) = (x, y, z, \psi, \theta = \alpha_0 - \alpha - \gamma, \phi)$, where $\alpha_0$ is the angle of attack at straight and level (zero pitch) flight. We assume that the state is fully observed (i.e. $\h(\x_t) = \x_t$) up to the measurement noise $\w_t$, and in our experiments we consider Gaussian noise distributions $\V^u_t$, $\V^x_t$, and $\W_t$. We note that the explicit consideration of control noise $\v^u_t$ ensures non-Gaussian uncertainty distributions at every time step, in addition to those arising from non-linear propagation.

\subsection{Performance of Adaptive Mixture Importance Sampling}\label{subsec:results}

\begin{figure}[t]
\centering
    \includegraphics[width=0.45\textwidth]{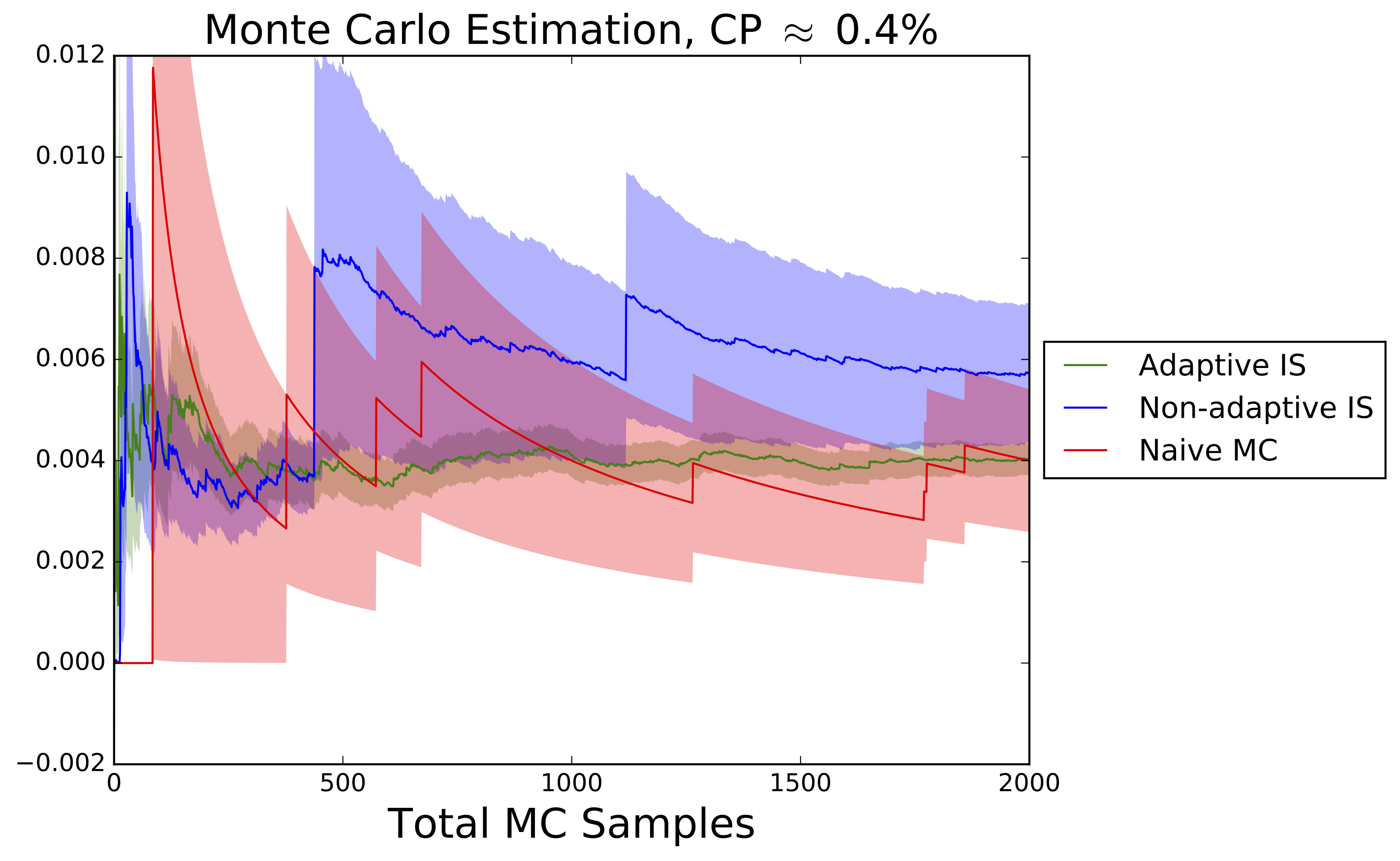}\\
    \includegraphics[width=0.45\textwidth]{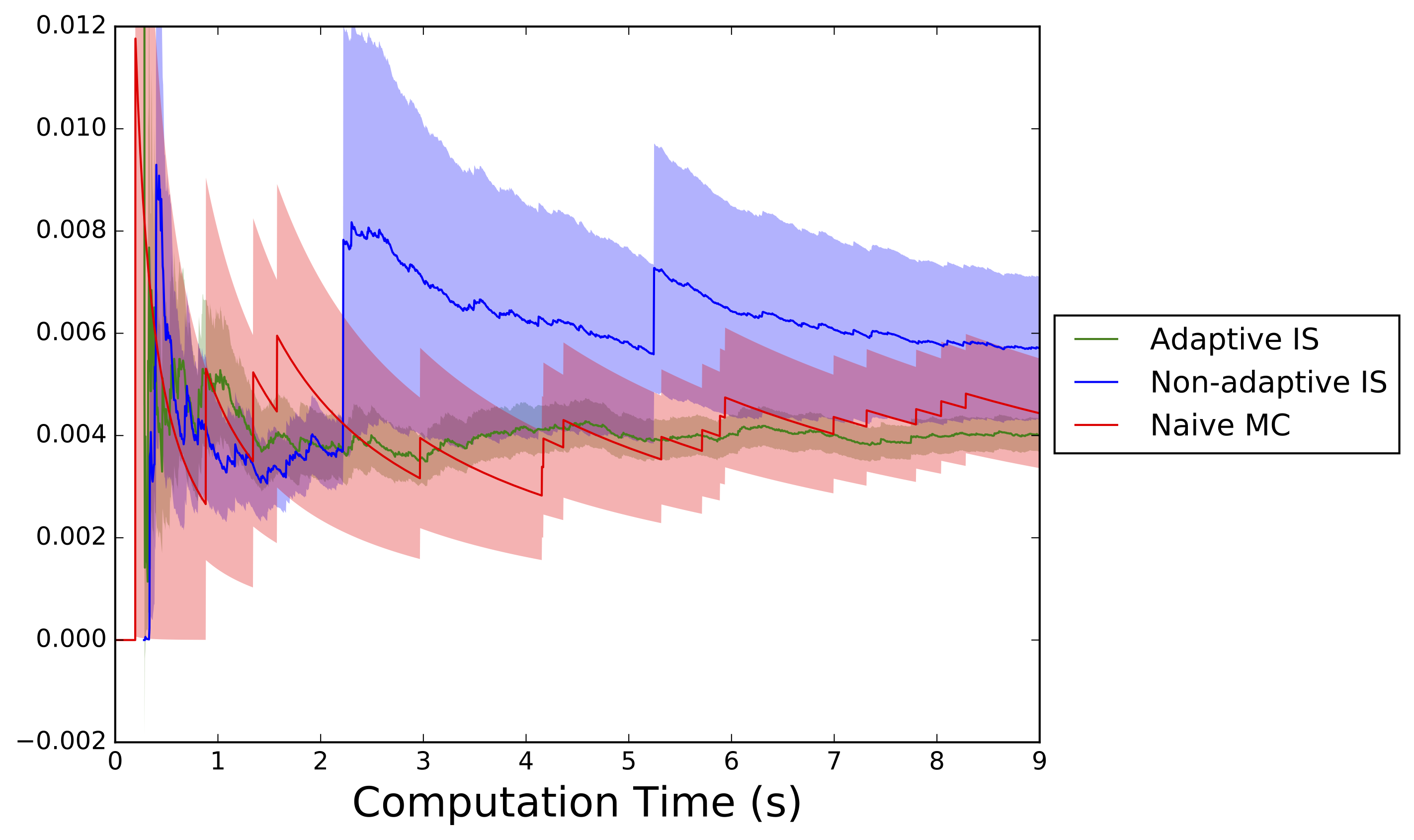}
    \vspace*{.3cm}\\
    {\small\begin{tabular}{| c | c | c | c | c | c |}
    \hline
    \multicolumn{6}{|c|}{Estimator means computed over 30 trials, $m = 1000$ samples}\\
    \hline
    $\hat p_{\text{AIS}}$ & $\hat p_{\text{IS}}$ & $\hat p_{\text{NMC}}$ & $\hat \sigma_{\text{AIS}}$ & $\hat \sigma_{\text{IS}}$ & $\hat \sigma_{\text{NMC}}$\\
    \hline
    0.436\% & 0.416\% & 0.471\% & 0.043\% & 0.048\% & 0.209\%\\
    \hline
    \end{tabular}}
        \caption{Example runs of Algorithm~\ref{alg:mixtureIS} with $k=20, \ell = 100$ (adaptive mixture IS), with $k=2000, \ell = 1$ (non-adaptive IS), and naive Monte Carlo for the nominal trajectory depicted in Figure~\ref{fig:plane_crashes}. The dark lines represent the evolution of the MC estimators $\hat p$; the shadows around each line represent a confidence interval of $\pm 1$ standard error, estimated as $\hat\sigma = \sqrt{\hat V}$. Both IS methods are shifted by 0.28 s in the lower plot to reflect the time required to derive the noise parameters $\eta$. Adaptive IS converges to an estimate with a usable level of certification within $500$ samples (2 s).}
    \label{fig:cp_evolution}
\end{figure}

\begin{table*}[htbp]
\centering
\caption{Mixture weights $\alpha$ derived through Algorithm~\ref{alg:mixtureIS} compared to half-space violation probability.\hspace{\textwidth}
Adaptive IS batch size $k=20$, defensive importance sampling weight $\alpha_{10}$ lower-bounded at $10\%$.}
\begin{tabular}{| l | c | c | c | c | c | c | c | c | c | c |}
\hline
Distribution component & $q_1$ & $q_2$ & $q_3$ & $q_4$ & $q_5$ & $q_6$ & $q_7$ & $q_8$ & $q_9$ & $q_{10}$ ($=P$) \\
\hline
Time step $t$ & 82 & 83 & 83 & 2 & 13 & 14 & 14 & 68 & 82 & N/A\\
\hline
Plane component & Wing & Stabilizer & Wing & Tail & Wing & Wing & Stabilizer & Wing & Body & N/A\\
\hline
Half-space violation probability & 0.177\% & 0.120\% & 0.111\% & 0.065\% & 0.041\% & 0.029\% & 0.026\% & 0.016\% & 0.012\% & N/A\\
\hline
Example $\alpha^{100}$ ($m=2000$)& 0.165 & 0.097 & 0.091 & 0.101 & 0.038 & 0.079 & 0.062 & 0.047 & 0.069 & 0.247\\
\hline
Example $\alpha^{1000}$ ($m=20000$)& 0.208 & 0.157 & 0.191 & 0.115 & 0.043 & 0.057 & 0.032 & 0.037 & 0.058 & 0.1\\
\hline
\end{tabular}\label{tab:weights}
\end{table*}

We implemented Algorithms~\ref{alg:mixtureIS} and \ref{alg:cop} in Julia \cite{BezansonKarpinskiEtAl2012} using the Bullet physics engine \cite{Coumans} for continuous (swept) collision checking, and ran experiments on a Linux system equipped with a 3.0GHz 8-core Intel i7-5960X processor (although we note that the implementation presented in this work is only single-threaded). Figure~\ref{fig:cp_evolution} depicts estimation results for applying adaptive mixture IS (Alg. 1, $k=20$, $\ell = 100$), non-adaptive IS (Alg. 1, $k=2000$, $\ell = 1$), and naive MC ($Q = P$ in Equations \eqref{eq:ISp} and \eqref{eq:ISV}) to the nominal trajectory depicted in Figure~\ref{fig:plane_crashes}. For discrete LQG control, $T = 100$ and the time discretization is $\Delta t = 0.129\text{ s}$. Both importance sampling methods use $D = 10$ mixture components, initialized with uniform weight aside from $q_{10} = P$ initialized with $\alpha^1_{10} = 0.5$. Figure~\ref{fig:cp_evolution} is indicative of a problem that may arise from poorly chosen mixture weights. Non-adaptive IS (in blue) twice encounters positive collision samples with a very high likelihood ratio, indicating poor proportional representation by the IS distribution $Q$. The table in Figure~\ref{fig:cp_evolution} indicates that this type of event is relatively rare, as on average $\hat \sigma_{\text{IS}}$ is much lower than $\hat \sigma_{\text{NMC}}$ (owing to the construction of a good $\eta$) and near adaptive mixture IS in terms of error certificate. 

On average both importance sampling methods process $1000$ samples in $\sim 4.5\text{ s}$ (specifically, AIS: $4.46\pm0.13\text{ s}$, IS: $4.51\pm0.65\text{ s}$). We can see from Figure~\ref{fig:cp_evolution}, however, that only a few hundred samples are required to get a confident handle on trajectory CP, which we note may be processed in far less time than the airplane takes to fly its 13 s trajectory. The equivalent timings for both versions of IS are not surprising for the single-threaded implementation featured in this work, as the computational effort required by the stochastic mirror descent update is negligible compared to integrating dynamics, collision checking, and computing likelihood ratios. A parallel implementation might require larger batch sizes $k$ in order to overcome the communication overhead inherent in coordinating adaptive IS and achieve the expected parallel MC speedup.
Naive MC processes $1000$ samples in $2.34\pm .04\text{ s}$; the additional importance sampling time is spent entirely in evaluating probability density functions.

Table~\ref{tab:weights} gives a description of the mixture components used in importance sampling, as well as the final adaptive IS mixture weights from Figure~\ref{fig:cp_evolution}. We can see from the last two rows that a few thousand samples is generally insufficient to converge to the optimal weights, but already at $m=2000$ samples the relationship between good weights and half-space violation probability for the approximate marginal distributions defies heuristic approximation.

\section{Conclusions} \label{sec:conc}
We have presented an adaptive mixture importance sampling algorithm inspired by the statistics literature \cite{Ryu2016}
and demonstrated its success in quantifying CP (with tight estimated error) for an LQG with EKF control policy applied to
a non-linear system with a full rigid-body collision model. In particular, we note that this procedure succeeds in achieving a level of certifiable accuracy
for which there are no comparable existing methods other than naive, non-variance-reduced, Monte Carlo. The adaptive nature of the procedure has been demonstrated as essential
for selecting proper component weights for use in mixture IS, at negligible additional computational cost compared to non-adaptive mixture IS.

While this work may in its current form see direct application within a non-linear LQG control planner (for which we stress that a parallel
implementation would be a key technology for enabling truly real-time use), we mention here a number of other future research avenues.
First, we note that in this work a few thousand samples are sufficient to learn improved mixture weights. With a budget of tens of thousands of MC samples (possibly enabled by GPU), we might attempt to adaptively improve the high-dimensional distribution parameters $\eta$ as well as the weights $\alpha$.
Second, estimating CP for control policies departing from the LQG approach of tracking a nominal trajectory (e.g., stochastic extended LQR \cite{SunBergEtAl2016}) may be considered using the same mixture IS techniques.
Third, extension to environment uncertainty in addition to robot dynamics uncertainty may similarly draw inspiration from CP approximation heuristics \cite{DuToitBurdick2011,ParkParkEtAl2016} to inform the construction of IS component distributions.
Finally, we note that although the ``rare event'' considered throughout this paper has been obstacle collision, adaptive importance
sampling as a Monte Carlo variance reduction technique may be applied to estimate a variety of other performance or safety requirements.

{
\section*{Acknowledgments}
This work was supported by a Qualcomm Innovation Fellowship, by the Toyota Research Institute (``TRI''), and by NASA under the Space Technology Research Grants Program, Grant NNX12AQ43G.
}

\bibliographystyle{plainnat}
\bibliography{../../../bib/main,ASL_papers}

\end{document}

%% file: preamble.tex
\usepackage{color}
\usepackage{amsmath}
\usepackage{amssymb}
\usepackage{graphicx}
\usepackage{comment,xspace}
\usepackage{hyperref}
\usepackage{fancybox}

\newcommand{\argmin}{\operatornamewithlimits{argmin}}

\newcommand{\E}[1]{\mbox{$\mathbb{E}\left[#1\right]$}} 
\newcommand{\EP}[1]{\mbox{$\mathbb{E}^P\left[#1\right]$}} 
\newcommand{\EQ}[1]{\mbox{$\mathbb{E}^Q\left[#1\right]$}}
 
\newcommand{\VarQ}[1]{\mbox{$\mathbf{Var}^Q\left[#1\right]$}}  
\newcommand{\X}{\mathbf{X}}
\newcommand{\R}{\mathbb{R}}
\newcommand*\diff{\mathop{}\!\mathrm{d}}

\newcommand{\dx}{\diff x}
\newcommand{\x}{\mathbf{x}}
\renewcommand{\u}{\mathbf{u}}
\renewcommand{\v}{\mathbf{v}}
\newcommand{\w}{\mathbf{w}}
\newcommand{\q}{\mathbf{q}}

\newcommand{\V}{\mathbf{V}}
\newcommand{\W}{\mathbf{W}}
\newcommand{\y}{\mathbf{y}}
\newcommand{\s}{\mathbf{s}}
\newcommand{\g}{\mathbf{g}}
\newcommand{\z}{\mathbf{z}}
\newcommand{\f}{\mathbf{f}}
\newcommand{\h}{\mathbf{h}}
\newcommand{\xb}{\bar{\mathbf{x}}}
\newcommand{\xh}{\hat{\mathbf{x}}}
\newcommand{\ub}{\bar{\mathbf{u}}}
\newcommand{\zb}{\bar{\mathbf{z}}}
\newcommand{\0}{\mathbf{0}}
\newcommand{\pp}{\mathbf{p}}
\newcommand{\Xobs}{\mathcal{X}_{\mathrm{obs}}}

%% file: Schmerling.Pavone.RSS17.ARXIV.bbl
\newcommand{\noopsort}[1]{} \newcommand{\printfirst}[2]{#1}
  \newcommand{\singleletter}[1]{#1} \newcommand{\switchargs}[2]{#2#1}
\begin{thebibliography}{26}
\providecommand{\natexlab}[1]{#1}
\providecommand{\url}[1]{\texttt{#1}}
\expandafter\ifx\csname urlstyle\endcsname\relax
  \providecommand{\doi}[1]{doi: #1}\else
  \providecommand{\doi}{doi: \begingroup \urlstyle{rm}\Url}\fi

\bibitem[ONR(2012)]{ONR2012}
Science and technology strategic plan, {C4ISR}.
\newblock Technical report, {Office of Naval Research}, 2012.
\newblock {Available} at
  \url{http://www.onr.navy.mil/Science-Technology/Departments/Code-31.aspx}.

\bibitem[Aoude et~al.(2013)Aoude, Luders, Joseph, Roy, and
  How]{AoudeLudersEtAl2013}
G.~S. Aoude, B.~D. Luders, J.~M. Joseph, N.~Roy, and J.~P. How.
\newblock Probabilistically safe motion planning to avoid dynamic obstacles
  with uncertain motion patterns.
\newblock \emph{{Autonomous Robots}}, 35\penalty0 (1):\penalty0 51--76, 2013.

\bibitem[Beard and McLain(2012)]{BeardMcLain2012}
R.~W. Beard and T.~W. McLain.
\newblock \emph{Small Unmanned Aircraft: Theory and Practice}.
\newblock {Princeton University Press}, 2012.

\bibitem[Bezanson et~al.(2012)Bezanson, Karpinski, Shah, and
  Edelman]{BezansonKarpinskiEtAl2012}
J.~Bezanson, S.~Karpinski, V.~B. Shah, and A.~Edelman.
\newblock Julia: A fast dynamic language for technical computing, 2012.
\newblock {Available} at \url{http://arxiv.org/abs/1209.5145}.

\bibitem[Carlone and Karaman(2017)]{CarloneKaraman2017}
L.~Carlone and S.~Karaman.
\newblock Attention and anticipation in fast visual--inertial navigation.
\newblock In \emph{{Proc.\ IEEE Conf.\ on Robotics and Automation}}, 2017.

\bibitem[Coumans()]{Coumans}
E.~Coumans.
\newblock Bullet physics.
\newblock {Available} at \url{http://bulletphysics.org/}.

\bibitem[Dahm(2010)]{Dahm2010}
W.~J.~A. Dahm.
\newblock \emph{Technology Horizons: A Vision for {Air} {Force} Science \&
  Technology During 2010--2030}.
\newblock {Air Force Research Institute}, 2010.

\bibitem[Du~Toit and Burdick(2011)]{DuToitBurdick2011}
N.~E. Du~Toit and J.~W. Burdick.
\newblock Probabilistic collision checking with chance constraints.
\newblock \emph{{IEEE Transactions on Robotics}}, 27\penalty0 (4):\penalty0
  809--815, 2011.

\bibitem[Guibas et~al.(2008)Guibas, Hsu, Kurniawati, and
  Rehman]{GuibasHsuEtAl2008}
L.~J. Guibas, D.~Hsu, H.~Kurniawati, and E.~Rehman.
\newblock Bounded uncertainty roadmaps for path planning.
\newblock In \emph{{Workshop on Algorithmic Foundations of Robotics}}, 2008.

\bibitem[Haykin(2001)]{Haykin2001}
S.~Haykin, editor.
\newblock \emph{Kalman Filtering and Neural Networks}.
\newblock {John Wiley \& Sons}, 2001.

\bibitem[Janson et~al.(2015)Janson, Schmerling, and
  Pavone]{JansonSchmerlingEtAl2015b}
L.~Janson, E.~Schmerling, and M.~Pavone.
\newblock {Monte} {Carlo} motion planning for robot trajectory optimization
  under uncertainty.
\newblock In \emph{{Int.\ Symp.\ on Robotics Research}}, 2015.

\bibitem[Kothari and Postlethwaite(2013)]{KothariPostlethwaite2013}
M.~Kothari and I.~Postlethwaite.
\newblock A probabilistically robust path planning algorithm for {UAVs} using
  rapidly-exploring random trees.
\newblock \emph{{Journal of Intelligent \& Robotic Systems}}, 71\penalty0
  (2):\penalty0 231--253, 2013.

\bibitem[LaValle(2006)]{LaValle2006}
S.~M. LaValle.
\newblock \emph{Planning Algorithms}.
\newblock {Cambridge University Press}, 2006.

\bibitem[Liu and Ang(2014)]{LiuAng2014}
W.~Liu and M.~H. Ang, Jr.
\newblock Incremental sampling-based algorithm for risk-aware planning under
  motion uncertainty.
\newblock In \emph{{Proc.\ IEEE Conf.\ on Robotics and Automation}}, 2014.

\bibitem[Luders et~al.(2010)Luders, Kothari, and How]{LudersKothariEtAl2010}
B.~D. Luders, M.~Kothari, and J.~P. How.
\newblock Chance constrained {RRT} for probabilistic robustness to
  environmental uncertainty.
\newblock In \emph{{AIAA Conf.\ on Guidance, Navigation and Control}}, 2010.

\bibitem[Luders et~al.(2013)Luders, Karaman, and How]{LudersKaramanEtAl2013}
B.~D. Luders, S.~Karaman, and J.~P. How.
\newblock Robust sampling-based motion planning with asymptotic optimality
  guarantees.
\newblock In \emph{{AIAA Conf.\ on Guidance, Navigation and Control}}, 2013.

\bibitem[Luo et~al.(2016)Luo, Bai, Hsu, and Lee]{LuoBaiEtAl2016}
Y.~Luo, H.~Bai, D.~Hsu, and W.~S. Lee.
\newblock Importance sampling for online planning under uncertainty.
\newblock In \emph{{Workshop on Algorithmic Foundations of Robotics}}, 2016.

\bibitem[Oldewurtel et~al.(2008)Oldewurtel, Jones, and
  Morari]{OldewurtelJonesEtAl2008}
F.~Oldewurtel, C.~N. Jones, and M.~Morari.
\newblock A tractable approximation of chance constrained stochastic {MPC}
  based on affine disturbance feedback.
\newblock In \emph{{Proc.\ IEEE Conf.\ on Decision and Control}}, 2008.

\bibitem[Owen(2013)]{Owen2013}
A.~B. Owen.
\newblock {Monte} {Carlo} theory, methods and examples, 2013.
\newblock {Available} at \url{http://statweb.stanford.edu/~owen/mc/}.

\bibitem[Pan et~al.(2011)Pan, Chitta, and Manocha]{PanChittaEtAl2011}
J.~Pan, S.~Chitta, and D.~Manocha.
\newblock Probabilistic collision detection between noisy point clouds using
  robust classification.
\newblock In \emph{{Int.\ Symp.\ on Robotics Research}}, 2011.

\bibitem[Park et~al.(2016)Park, Park, and Manocha]{ParkParkEtAl2016}
C.~Park, J.~S. Park, and D.~Manocha.
\newblock Fast and bounded probabilistic collision detection in dynamic
  environments for high-{DOF} trajectory planning.
\newblock In \emph{{Workshop on Algorithmic Foundations of Robotics}}, 2016.

\bibitem[Patil et~al.(2012)Patil, van~den Berg, and
  Alterovitz]{PatilBergEtAl2012}
S.~Patil, J.~van~den Berg, and R.~Alterovitz.
\newblock Estimating probability of collision for safe motion planning under
  {Gaussian} motion and sensing uncertainty.
\newblock In \emph{{Proc.\ IEEE Conf.\ on Robotics and Automation}}, 2012.

\bibitem[Ryu(2016)]{Ryu2016}
E.~Ryu.
\newblock \emph{Convex Optimization for {Monte} {Carlo}: Stochastic
  Optimization for Importance Sampling}.
\newblock PhD thesis, {Stanford University}, 2016.

\bibitem[Sun et~al.(2013)Sun, Torres, van~den Berg, and
  Alterovitz]{SunTorresEtAl2013}
W.~Sun, L.~G. Torres, J.~van~den Berg, and R.~Alterovitz.
\newblock Safe motion planning for imprecise robotic manipulators by minimizing
  probability of collision.
\newblock In \emph{{Int.\ Symp.\ on Robotics Research}}, 2013.

\bibitem[Sun et~al.(2016)Sun, van~den Berg, and Alterovitz]{SunBergEtAl2016}
W.~Sun, J.~van~den Berg, and R.~Alterovitz.
\newblock Stochastic extended {LQR} for optimization-based motion planning
  under uncertainty.
\newblock \emph{{IEEE Transactions on Automation Sciences and Engineering}},
  13\penalty0 (2):\penalty0 437--447, 2016.

\bibitem[van~den Berg et~al.(2011)van~den Berg, Abbeel, and
  Goldberg]{BergAbbeelEtAl2011}
J.~van~den Berg, P.~Abbeel, and K.~Goldberg.
\newblock {LQG-MP}: Optimized path planning for robots with motion uncertainty
  and imperfect state information.
\newblock \emph{{Int.\ Journal of Robotics Research}}, 30\penalty0
  (7):\penalty0 895--913, 2011.

\end{thebibliography}
